\documentclass[10pt,twocolumn,letterpaper]{article}

\usepackage{iccv}
\usepackage{times}
\usepackage{epsfig}
\usepackage{graphicx}
\usepackage{amsmath}
\usepackage{amssymb}

\usepackage{times}
\usepackage{epsfig}
\usepackage{graphicx}
\usepackage{amsmath}
\usepackage{amssymb}
\usepackage{multirow}
\usepackage{booktabs}
\usepackage{mathtools}

\usepackage{caption}
\usepackage{subcaption}
\usepackage{algpseudocode,algorithm,algorithmicx}
\usepackage{xr}
\usepackage{bm}
\usepackage{color}
\usepackage[accsupp]{axessibility}
\usepackage{xcolor}

\usepackage[pagebackref=true,breaklinks=true,letterpaper=true,colorlinks,bookmarks=false]{hyperref}

\iccvfinalcopy 

\def\iccvPaperID{9521} 

\ificcvfinal\fi

\begin{document}

\title{Fg-T2M: Fine-Grained Text-Driven Human Motion Generation via Diffusion Model}


\author{Yin Wang$^{1}$,
Zhiying Leng$^{1,2}$,
Frederick W. B. Li$^{3}$,
Shun-Cheng Wu$^{2}$, 
Xiaohui Liang$^{1,4,}$\thanks{Corresponding author.} \\ [2mm]
\small$^{1}$State Key Laboratory of Virtual Reality Technology and Systems, Beihang University, Beijing, China\\
\small$^{2}$Technical University of Munich, Germany\\
\small$^{3}$Department of Computer Science, University of Durham, U.K.\\
\small$^{4}$Zhongguancun Laboratory, Beijing, China.
\\
{\tt\small \{wang\_yin,zhiyingleng,liang\_xiaohui\}@buaa.edu.cn, frederick.li@durham.ac.uk, shuncheng.wu@tum.de}
}

\maketitle
\ificcvfinal\thispagestyle{empty}\fi

\begin{abstract}

   Text-driven human motion generation in computer vision is both significant and challenging. However, current methods are limited to producing either deterministic or imprecise motion sequences, failing to effectively control the temporal and spatial relationships required to conform to a given text description. In this work, we propose a fine-grained method for generating high-quality, conditional human motion sequences supporting precise text description. Our approach consists of two key components: 1) a linguistics-structure assisted module that constructs accurate and complete language feature to fully utilize text information; and 2) a context-aware progressive reasoning module that learns neighborhood and overall semantic linguistics features from shallow and deep graph neural networks to achieve a multi-step inference. Experiments show that our approach outperforms text-driven motion generation methods on HumanML3D and KIT test sets and generates better visually confirmed motion to the text conditions.

\end{abstract}



\section{Introduction}

Human motion generation is a crucial task in computer vision with various applications in animation production, gaming, robot control, and movie script visualization. Obtaining human motion sequences through traditional software is a labor-intensive and tedious process, while motion capture is complex and expensive. Recently, with the advancements in deep learning and computer vision, learning-based human motion generation has emerged as a solution to this problem, leading to the development of associated generation methods based on multimodal data. The input multimodal data include music~\cite{kao2020temporally,li2021ai,ren2020self,starke2022deepphase}, motion categories~\cite{cervantes2022implicit,guo2020action2motion,petrovich2021action}, text~\cite{ahuja2019language2pose,bhattacharya2021text2gestures,ghosh2021synthesis,guo2022generating,guo2022tm2t,lin:vigil18,petrovich2022temos,plappert2018learning,tevet2022human,zhang2022motiondiffuse}, among others. Text-driven human motion generation has been a popular research topic, because of its convenience and human-friendliness. In particular, natural language comprises nouns, verbs, adverbs, etc. 
The mutual connections among different words in a sentence establish its semantics. Verbs define the action's category, while adverbs control the fineness of the action.
The interaction between words in syntax plays a vital role in determining the structure and meaning of a sentence. Failure to fully incorporate these text features may result in inadequate text modeling, causing the generated motion sequence to deviate from the intended meaning of the original text.
\begin{figure}[t]
  \centering
  \includegraphics[width=\linewidth]{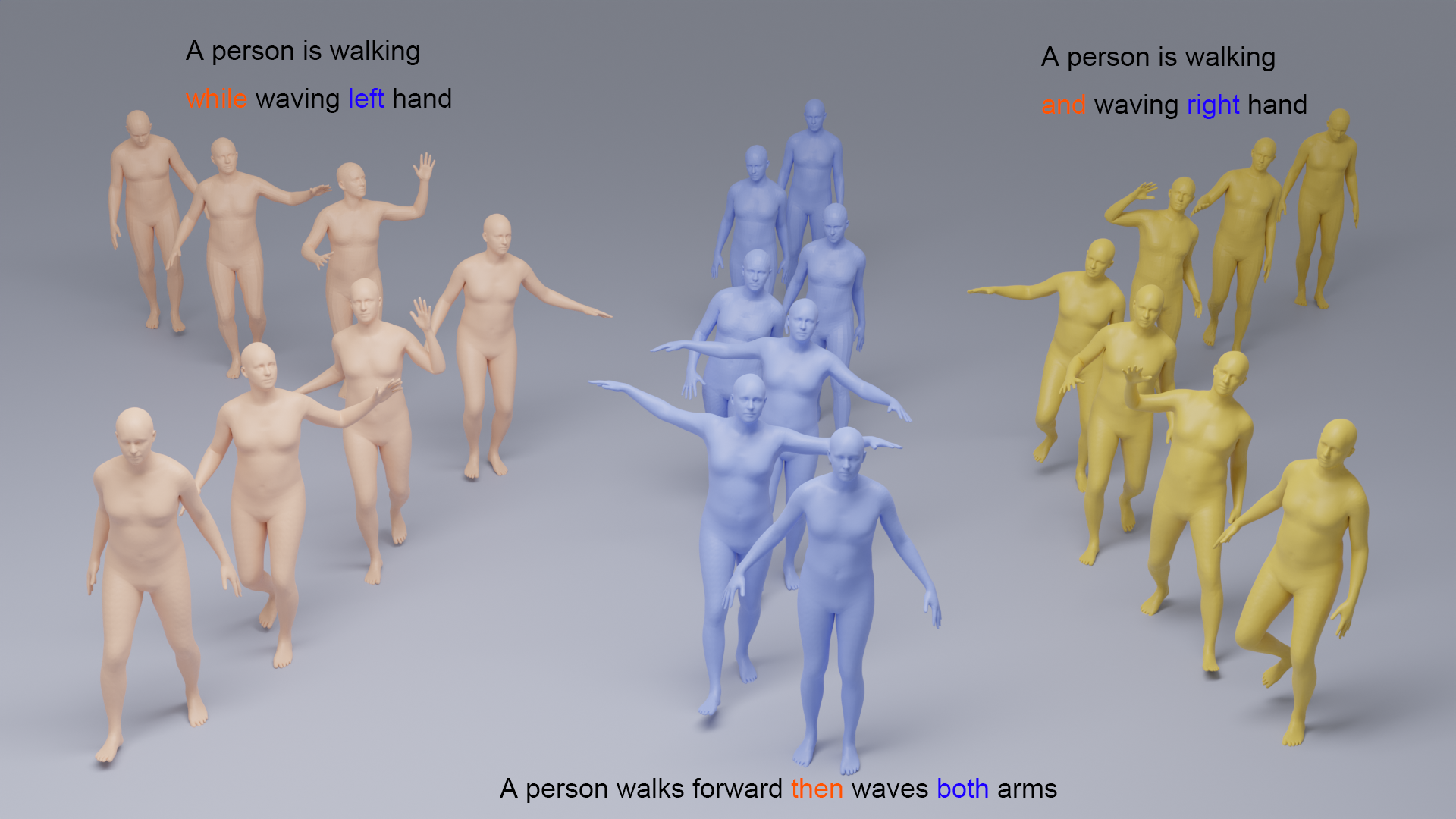}
  \caption{Our approach generates human motion sequences that grasp fine-grained details.}
\label{fig:fpic}
\end{figure}

Existing methods can be divided into two branches, including 1) cross-modal alignment of motion and text~\cite{ahuja2019language2pose,bhattacharya2021text2gestures,ghosh2021synthesis,guo2022generating,guo2022tm2t,lin:vigil18,petrovich2022temos}; 2) conditional diffusion models~\cite{tevet2022human,zhang2022motiondiffuse}. 
In the first methods, text sequences and motion sequences are mapped onto separate feature spaces and forcibly aligned, leading to a loss of original information from both domains. In the second methods, the diffusion model incorporates text information as a conditioning factor to learn the probability mapping of human motions. However, the model interacts with only one text feature at each time step of the inference process, lacking a progressive approach. Moreover, text modeling only involves simple manipulation, which ignores the importance of certain fine-grained words and leads to incomplete semantic understanding, making it challenging to learn focus points at each step. Overall, existing methods only use text information to a limited extent, which in turn affects the accuracy of motion generation based on the corresponding text content, especially for the motion in which texts contain fine-grained words. 
For instance, comprehending the sentence ``A man is walking forward while waving his right hand'' can be a difficult task, and expecting the model to grasp the fine-grained meaning of the terms ``while'' and ``right hand'' is even more demanding.


To tackle the aforementioned issues, we propose a fine-grained text-driven method for generating human motion sequences that precisely align with text prompts in Figure \ref{fig:fpic}.
Typically, people initially read a sentence to gain an overall semantic understanding before focusing on the fine-grained details of individual words. To replicate this process,  our method includes a linguistics-structure assisted module and a context-aware progressive reasoning module to fully utilize text information. Firstly, we utilize linguistics structure to facilitate information exchange between each text word. We use dependency parsing~\cite{nivre2008algorithms} to analyze the relationships among words in each sentence and construct a dependency tree, allowing each node to effectively communicate based on its dependent nodes and relationships. Then, the dependency tree nodes are passed to multi-layer graph neural networks to learn information aggregation. The multi-layer graph neural networks allow shallow network to learn neighborhood features as it can comprehend nearby details, and allow deep network to grasp overall semantic features because it is capable of aggregating information from entire nodes. Additionally, our GAT captures rich inter-word relationships while preserving the text linguistics structure by designing adaptive weights for each part-of-speech and dependency relation due to their distinctive role in the Text to Motion (T2M) task. 

Secondly, achieving the purpose of fine-grained interaction, context-aware progressive reasoning module performs a multi-step inference process with the progressive fusion of global and local information between text and motion, which is unprecedented in the T2M task.  This involves utilizing hierarchical semantic features to simulate the way humans comprehend sentences. We adopt the diffusion model framework and stack the hierarchical semantic features obtained from deep to shallow networks at each step to capture high-order relationships at different semantic levels. 
We evaluate our method  on HumanML3D dataset~\cite{guo2022generating} and KIT dataset~\cite{plappert2016kit}. Experiments show that  our approach outperforms the state-of-the-art methods and generates better visual motion.
Our main contributions include:
\begin{itemize}
\item To the best of our knowledge, we are the first to bring NLP methods into T2M task. Utilizing the structured understanding of the natural language prompts to help T2M models achieve better reasoning skills, which brings new ideas for the text-to-X community from a textual perspective.
\item We propose the Linguistics-Structure Assisted Module (LSAM), which utilizes a dependency parsing tree and graph networks to facilitate effective information exchange and data aggregation. It can obtain both neighborhood and overall semantic linguistic features.
\item We propose a Context-Aware Progressive Reasoning Module (CAPR) that implements a multi-step progressive inference strategy within the diffusion model framework, mimicking the human reading process by moving from global to local relationships.
\item Experimental results demonstrate that our proposed method outperforms previous methods, and achieves competitive performance on the HumanML3D and KIT datasets.
\end{itemize}

\section{Related Work}

\subsection{Motion generation model}
Generative models play a crucial role in motion synthesis by generating high-quality human motion. Generative Adversarial Nets (GAN)~\cite{goodfellow2020generative} use two sub-models: a generator model that produces new samples, and a discriminator model that attempts to classify samples as either real or fake. These two models compete against each other during training. However, the interpretability of GAN is poor because the learned data distribution lacks an explicit expression, resembling a black box mapping function.


Auto-Encoding Variational Bayes (VAE)~\cite{kingma2013auto} is a widely used generative model in motion synthesis. Its primary objective is to generate new samples from the learned distribution of objects by learning latent attributes from the probability distribution of the latent variable space, thereby constructing new examples. Despite its usefulness, the quality of samples generated by VAE can be improved.

Recently proposed diffusion models~\cite{ho2020denoising,nichol2021improved,song2020denoising} have shown immense potential in modeling and present an exciting opportunity to expand into text-driven motion generation. These models utilize the stochastic diffusion process modeled in thermodynamics, which gradually adds noise to the samples of the data distribution. The deep learning model then learns the reverse process of denoising the samples gradually. Diffusion models have the advantage over previous models as they do not make any assumptions about the target distribution, leading to a more diverse generation and better suitability for our task. Therefore, we propose a novel fine-grained human motion generation method that employs the Denoising Diffusion Probability Model~\cite{ho2020denoising}.


\begin{figure*}[t]
    \centering
     \includegraphics[width=\linewidth]{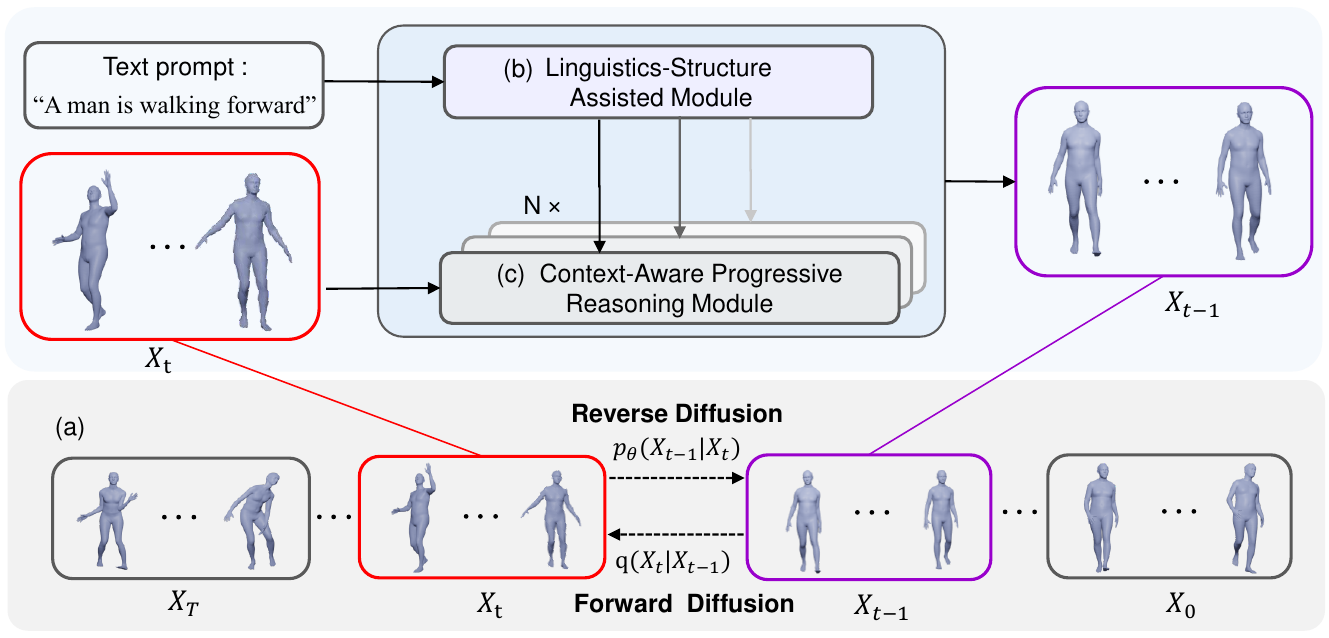}
    \caption{\textbf{Overall Pipeline of our proposed method Fg-T2M}: (a) the reverse denoising process of the diffusion model from $X_T$ to $X_0$; (b) the text encoder with proposed linguistics-structure assisted module with dependency parsing in section~\ref{sec:LSAM}; (c) the motion decoder with introduced context-aware progressive reasoning module in section ~\ref{sec:CAPR}.}
    \label{mypipeline}
\end{figure*}
\subsection{Text-driven human motion generation}
The task of text-driven human motion generation involves generating 3D human motion sequences that conform to a textual description. Several previous works have tackled this task. Initially, Text2Action~\cite{ahn2018text2action} proposed short-text conditioned motion generation based on an RNN model. Subsequently, Ahuja \etal~\cite{ahuja2019language2pose} and Ghosh \etal \cite{ghosh2021synthesis} focused on creating a joint representation of text and motion by projecting both features into a shared latent space. However, these methods involve a one-to-one mapping between text and motion, implying that given the same text, they can only produce fixed motion sequence results.

To increase generated result diversity, TEMOS~\cite{petrovich2022temos} introduced a VAE architecture that finds a joint latent space for motion and text under Gaussian distribution constraints. Guo \etal~\cite{guo2022generating} used a temporal VAE to autoregressively generate motion sequences based on text features. However, these methods have a significant drawback of mapping text and motion sequences to separate feature spaces and forcibly align them, leading to a loss of information in both domains. Recently, diffusion models have shown great potential in image generation and have inspired the development of diffusion models for human motion generation. Tevet \etal~\cite{tevet2022human} and Zhang \etal~\cite{zhang2022motiondiffuse} encoded text descriptions using pre-trained models and estimated Gaussian noise or the original motion sequence at each reverse diffusion step. However, their text modeling is often crude and does not fully leverage linguistic structure for sentence semantics. Also, they do not have a progressive process that allows the diffusion model to focus on different content at different time stamps, and they only interact with fixed text features during the inference process.

\section{Preliminaries}

The diffusion process is a Markov process consisting of a forward process and a reverse process.
The forward process starts with the real data $X_0$ at step 0 and proceeds in a Markovian manner by adding Gaussian noise at each step. Over $t$ steps, $X_0$ is transformed into $X_t$, which is close to the Gaussian distribution $N(0, I)$. As a result, the original motion sequence is converted into a complete noise distribution, which can be expressed as:
\begin{equation}
        q(\mathbf{x}_t \vert \mathbf{x}_{t-1}) \,=\, \mathcal{N}(\mathbf{x}_t; \sqrt{1-\beta_t}\mathbf{x}_{t-1}, \beta_t I)
\end{equation}
where $\beta_t$ is a hyper-parameter that controls the diffusion rate. The entire diffusion forward process is formulated as:
\begin{equation}
        q(\mathbf{x}_{1:T} \vert \mathbf{x}_0) \,=\, \prod_{t=1}^{T} q(\mathbf{x}_t \vert \mathbf{x}_{t-1})
\end{equation}
where $T$ denotes total steps in diffusion. The diffusion reverse process samples from the Gaussian distribution $X_t$ as the initial input and attempts to gradually remove the noise on a reverse Markov-chain, which  can be defined as follows:
\begin{equation}
        p_{\theta}(\mathbf{x}_{0:T}) \,=\, p(\mathbf{x}_{T}) \prod_{t=1}^{T}p_{\theta}(\mathbf{x}_{t-1} \vert \mathbf{x}_{t})
\end{equation}
\begin{equation}
        p_{\theta}(\mathbf{x}_{t-1} \vert \mathbf{x}_{t}) \,=\, \mathcal{N}(\mathbf{x}_{t-1}; \mu_\theta(\mathbf{x}_{t}, t), \Sigma_\theta(\mathbf{x}_{t}, t))
\end{equation}
where $\mu_\theta$ is the estimated item by model, $t$ is the timestep indicating where the denoising process has conducted.
\section{Method}

To enhance the modeling of fine-grained human motion generation, we present our proposed method in Figure \ref{mypipeline}. Previous methods~\cite{ahn2018text2action,ahuja2019language2pose,ghosh2021synthesis} have adopted a coarse approach to text modeling, leading to an underutilization of text information. Moreover, they treat all words equally without considering their importance and uniqueness. In contrast, we leverage linguistic structures in sentences to further enhance text encoding and differentiate between overall semantic information and detailed features using a multi-step progressive reasoning strategy.

Given a text prompt, $W = \{w_1,w_2,\dots,w_n\} $, $W  \in \mathbb{R}^{N \times L}$ where $N$ represents the number of words and  $L$ is the dimension of word vector. 
Our goal is to generate a human motion sequence, denoted as $M = \{m_1,m_2,\dots m_t\}$, where $M \in \mathbb{R}^{T \times D}$. Here, $T$ refers to the sequence length and $D$ is the motion representation dimension. To achieve this, we introduce Fg-T2M, a method that generates motion sequences that align well with the corresponding textual content. In the following, we provide an overview of our approach in section~\ref{sec:architecture}, followed by the introduction of the Linguistics-Structure Assisted Module (LSAM) in section~\ref{sec:LSAM}. Lastly, we present the Context-Aware Progressive Reasoning (CAPR) module in section~\ref{sec:CAPR}.


\subsection{Overview}
\label{sec:architecture}


Figure \ref{mypipeline} shows our pipeline for generating motion sequences. We randomly sample $\mathbf{x}_T$ from distribution $N(0,I)$, input $\mathbf{x}_T$, current step $T$, and text control condition $c$ to obtain $\mathbf{x}_{T-1}$, iterating $T$ rounds until we get $\mathbf{x}_0$. In the denoising process, the condition $c$ is fed into our LSAM text encoder. By leveraging dependency relationships between words, we use graph neural networks to aggregate data and extract hierarchical semantic features. The motion decoder is stacked with CAPR. Multiple layers of these modules enhance the quality of the generated motion sequence.

To train the motion diffusion model, we optimize the objective to predict the original data, representing as follows:
\begin{equation}
    \mathcal{L}=\mathrm{E}_{t \in [1,T], \mathbf{x}_0 \sim q(\mathbf{x}_0),\epsilon \sim \mathcal{N}(0, I)} [\parallel \mathbf{x}_0 - \hat{\mathbf{x}}_0(\mathbf{x}_t,t,c) \parallel_2^2].
\end{equation}
The regular L2 loss can improve performance for all geometric losses mentioned above.


\subsection{Linguistics-Structure Assisted Module}
\label{sec:LSAM}

\begin{figure}[t]
  \centering
  \includegraphics[width=0.9\linewidth]{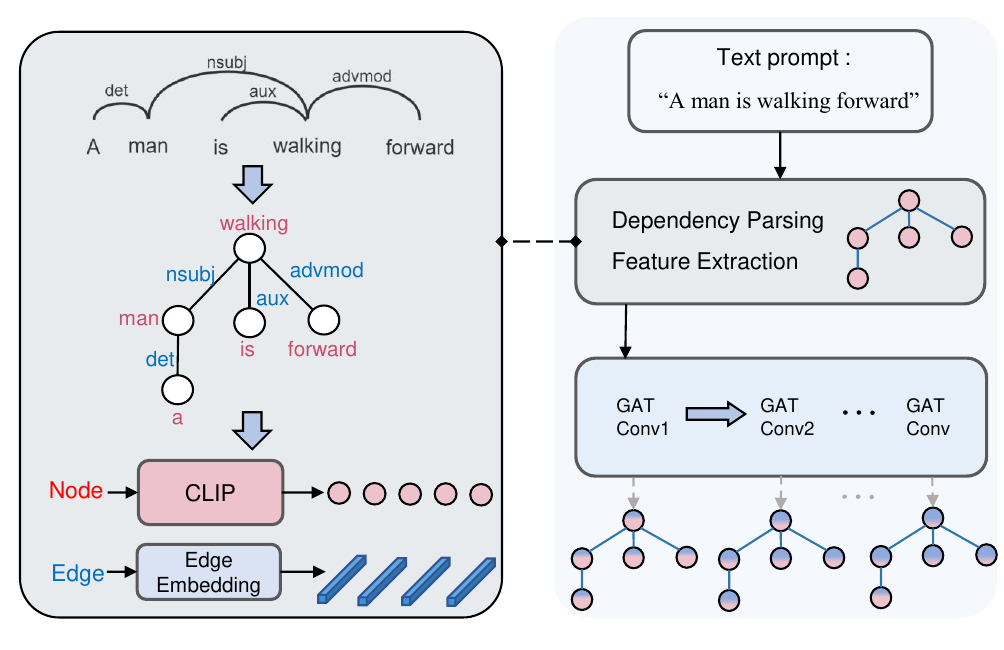}
  \caption{\textbf{Architecture of LSAM.}
  The left panel illustrates the process of obtaining node and edge features for one sentence in dependency analysis feature extraction, while the right panel demonstrates the workflows that utilize multiple layers of Graph Attention Network to extract feature information with varying degrees.}
\label{fig:LSAM}
\end{figure}

The current text modeling method is limited in its sensitivity to fine-grained words, which hinders the effectiveness of subsequent generations. To address this, we enhance the context modeling of sentences by identifying the phrase structure and syntactic relationships between phrases using dependency parsing, as shown in Figure \ref{fig:LSAM}. Linguistic structure, which comprises phrase structure and syntax relationships, helps us better understand the essence of text and the differences and similarities between sentences. Dependency parsing~\cite{nivre2008algorithms} precisely analyzes vocabulary and syntax to identify the dependency relationships between words in a sentence. In this approach, each word is treated as a node, and the dependency relationships between words are represented by edges that indicate syntactic connections. The resulting node and edge features representing the dependency relationships are input into a graph network to obtain multi-level semantics.
In summary, given a sentence, the dependency parsing analyses to build a text tree, nodes for words, and links between nodes for their linguistic relationships. The text tree structure as the graph initialization for graph topology and edge features provides GAT a better prior. The GAT aims to capture inter-word relationships while maintaining the linguistic structure of the text.

To extract word associations, we parse the dependency of phrase and obtain the hierarchical syntactic relationships using Spacy\footnote{Spacy: \url{https://spacy.io/}} for dependency parsing, as shown in the left panel of Figure \ref{fig:LSAM}. Spacy is a natural language processing software library for text processing, including lexical analysis, syntactic analysis, and more. In dependency parsing, each word is treated as a node, and the edges represent the labels of the dependency relationships between words, allowing us to construct a tree of language structure for the given text. We extract features using a graph attention network (GAT)~\cite{velivckovic2017graph}. For the input nodes of the graph network, we use word features obtained from the CLIP~\cite{radford2021learning} model. For the adjacency matrix, we set a value of 1 or 0 to represent the presence or absence of a dependency relationship between nodes. Since the number of dependency relationships is fixed, we use one-hot encoding to obtain a one-hot label for each relationship, which is then fed into an embedding layer. This can be described as:
\begin{equation}
    \mathbf{v_e}=\beta_i(\varphi_e(\operatorname{F_{onehot}}(R_i)))
\end{equation}
where $\mathbf{v_e} \in \mathbb{R}^{D_e}$, $D_e$ is the dimension of edge feature, $\varphi_e$ denotes an embedding layer, $\beta_i$ is adaptive weight parameters to be learned for every $R_i$, while $R_i$ indicates the edge relation between two nodes.
Since each word may be adjacent to multiple nodes, and different nodes contribute differently to semantics, it is necessary to distinguish them during the text feature extraction process.
We achieve this by applying GAT~\cite{velivckovic2017graph}, which extracts multi-level features from the obtained node information, adjacency matrix, and edge information. GAT~\cite{velivckovic2017graph} considers differences between nodes during information aggregation and handles irregularities between different nodes effectively, which is described as:
\begin{equation}
{x}_{i}^{\prime}=\alpha_{i, i}  \Theta {x}_{i} +\sum_{j \in N(i)} \alpha_{i, j} \Theta {x}_{j}
\end{equation}
\begin{equation}
    \alpha_{i, j} =\frac{\operatorname{exp} (\operatorname{F(\omega^{\mathsf{T}} [ \Theta {x}_{i}||\Theta {x}_{j}||\Theta_{e} {e}_{i, j} ])} ) }
    {\sum_{k \in N(i)\cup {\{i\}}} \operatorname{exp} (\operatorname{F}  (\omega^{\mathsf{T}} [\Theta {x}_{i}||\Theta {x}_{k}||\Theta_{e} {e}_{i, k}] )  ) }
\end{equation}

We have $x, \alpha \in \mathbb{R}^{N \times L}$, where $\alpha_{i, j}$ represents attention coefficients, F denotes $\operatorname{LeakyReLU}$, $e$ is edge features, $\omega$ and $\Theta$ represents the weight parameters to be learned, and $||$ is the concatenation operation. GAT~\cite{velivckovic2017graph} can model high-order dependency relationships by stacking multiple graph attention layers to capture global and local graph topology effectively. For example, stacking three layers of GAT~\cite{velivckovic2017graph} results in $x_i^1$, $x_i^2$, and $x_i^3$ for each node $x_i$, as follows:
\begin{equation}
{x}_{i}^{l+1}=\alpha_{i, i}^{l}  \Theta {x}_{i}^{l} + \sum_{j \in N(i)} \alpha_{i, j}^{l} \Theta {x}_{j}^{l} 
\end{equation}


The shallow network $x_i^1$ learns detailed features of the neighborhood as it can only gather information from directly adjacent nodes in the first step of node aggregation. This process captures more local and fine-grained information. By comparison, the deep network $x_i^3$ can learn overall semantic features by aggregating information from distant nodes after multiple aggregation steps. This enables it to capture more global and holistic information.

\subsection{Context-Aware Progressive Reasoning Module}
\label{sec:CAPR}
\begin{figure}[t]
  \centering
  \includegraphics[width=0.9\linewidth]{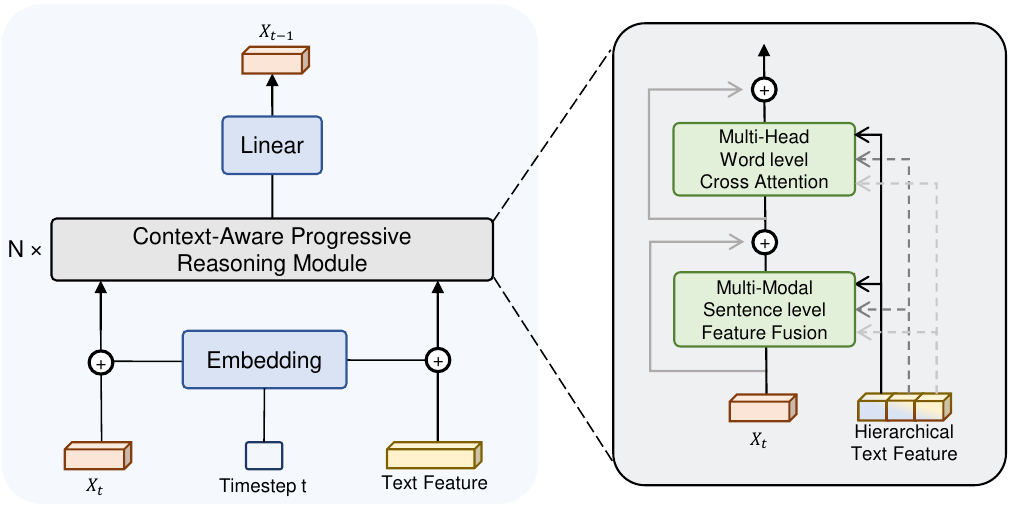}
  \caption{\textbf{Architecture of CAPR.} The left panel illustrates inferring $X_{t-1}$ from $X_t$, $t$, and text features. The right panel provides the detailed view of one context-aware progressive reasoning module, which comprises two parts: sentence-level feature fusion and word-level cross-attention.}
  \label{fig:CAPR}
\end{figure}


Guided by the Linguistics-Structure Assisted Module (LSAM) in the previous section, we stack Context-Aware Progressive Reasoning Modules to perform multi-step progressive reasoning in a structured manner. Unlike previous methods that learn fixed features~\cite{tevet2022human,zhang2022motiondiffuse}, our stacked modules grasp features from global to local. Each block receives distinct contextual information, with higher-level blocks utilizing deeper context features from the LSAM, and lower-level blocks utilizing shallower features. This results in the model perceiving hierarchical information during the inference process, which greatly benefits its ability to hierarchically comprehend the meaning of text content and sense its fine-grained words.

The CAPR comprises two parts: Multi-Modal Sentence-level Feature-Fusion and Multi-Head Word-level Cross-Attention. Figure \ref{fig:CAPR} shows that we start with the motion feature $x_t\in \mathbb{R}^{T \times D}$, text feature $W \in \mathbb{R}^{N \times L}$, and timestep $t$. 
To better capture the unique characteristics of each time step in the diffusion model, we first perform sinusoidal time embedding through linear layers to obtain timestep embedding $emb_t$. We then add the motion and text features to $emb_t$ to incorporate different time information at each timestep.


\textbf{Multi-Modal Sentence-Level Feature-Fusion} module fuses sentence-level text features and motions to obtain multi-modal features, as shown in Figure \ref{fig:compute}. For the i-th Context-Aware Progressive Reasoning Module block, the text $W_i$ is transformed to a sentence-level feature $S$ using convolution, which is derived from the i-th LSAM layer:

  \begin{figure}[h]
  \centering
  \includegraphics[width=0.9\linewidth]{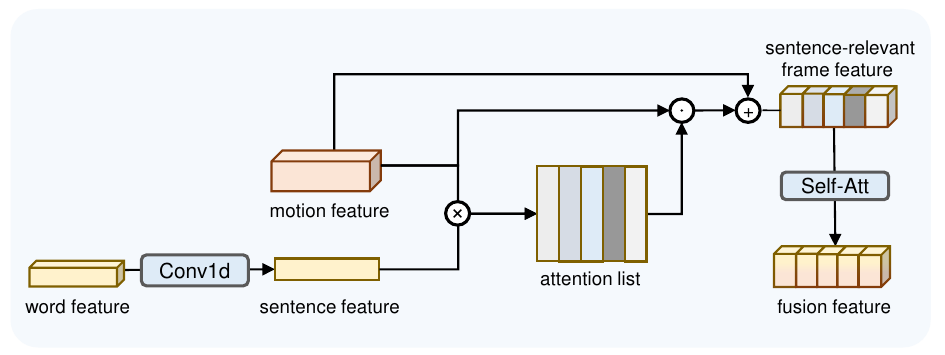}
  \caption{Illustration of \textbf{Multi-Modal Sentence-Level Feature-Fusion Module}.}
  \label{fig:compute}
\end{figure}

\begin{equation}
    S_i = \operatorname{Conv_{1d}(W_i)}
\end{equation}
where $S_i \in \mathbb{R}^{1 \times D_s}$, $W_i \in \mathbb{R}^{N \times D_w}$, $D_s$ and $D_w$ denotes the dimension of sentence feature and words feature, and $Conv_{1d}$ is the 1d convolution. 
We then perform a matrix multiplication to obtain an attention list $A$:
\begin{equation}
    A =\text{X}_t S_i^\mathrm{T}
\end{equation}
where $\text{X}_t \in \mathbb{R}^{T \times D}$ and $A \in \mathbb{R}^{T \times 1}$. 
Here, $\text{X}_t$ represents the motion feature at timestep $t$, and $A$ calculates the feature relevance between the sentence and each frame on the feature map. The resulting cross-modal feature, $\text{X}_t^{\prime}$, highlights the sentence-relevant frame feature channels:
\begin{equation}
    \text{X}_t^{\prime} = \text{X}_t + \lambda (\text{X}_t \odot \operatorname (\sigma(A)))
\end{equation}
where $\lambda$ is a hyper-parameter, $\odot $ is element-wise multiplication and $\sigma$ is a sigmoid activation function. 
To enhance modeling the correlation between different fused information, a self-attention mechanism~\cite{vaswani2017attention} is added to strengthen the connection between multiple frames. Multi-head self-attention is conducted on the fusion feature $\text{X}_t^{\prime}$ as follows:
\begin{equation}
\begin{aligned}
    \text{Q} = W_q\,  \text{X}_t^{\prime},\;\;
    \text{K} = W_k\,  \text{X}_t^{\prime},\; \operatorname{and}\;
    \text{V} = W_v\,  \text{X}_t^{\prime} 
\end{aligned}
\end{equation}
where $W_q$, $W_k$ and $W_v$ are trainable weights to generate \text{Q}, \text{K} and \text{V}, respectively. Obtain the attention scores using the formula below, where $\otimes$ is the matrix multiplication.
\begin{equation}
\begin{aligned}
    \operatorname{Attention(\text{Q, K, V})}=\operatorname{softmax}(\frac{\text{Q}\otimes \text{K}^{\top}}{\sqrt{d}} )\text{V}
\end{aligned}
    \label{attention}
\end{equation}



\textbf{Multi-Head Word-Level Cross-Attention} module learns the cross-interaction between motion sequences and multi-level contextual text features. It uses the formulas mentioned in the self-attention module for calculation, but modifies the motion features of K and V to text features. Therefore, on the stacked Context-Aware Progressive Reasoning modules $B_1, B_2, ..., B_i$, Block $B_i$ utilizes different word-level semantic features by:
\begin{equation}
    \hat{\text{X}}_t = \text{X}_t^{\prime} + \operatorname{Attention}(\text{X}_t^{\prime}  , \text{W}_i, \text{W}_i )
\end{equation}
where $\text{X}_t^{\prime}$ is motion features and $\text{W}_i$ is the text features. $\text{X}_t^{\prime}$ forms the query vector $Q$, $\text{W}_i$ forms the key vector $K$ and value vector $V$. They pass through the process in Equation~\ref{attention}. Finally, several MLP layers further processes the above features to output the predicted target $\text{X}_t \in \mathbb{R}^{T \times D}$.


\section{Experiments}

\begin{table}[t]
\centering
\resizebox{\columnwidth}{!}{%
\begin{tabular}{@{}lccccccc@{}}
\toprule
\multirow{2}{*}{Methods} & \multirow{2}{*}{R-TOP1 $\uparrow$} & \multirow{2}{*}{R-TOP2 $\uparrow$} & \multirow{2}{*}{R-TOP3 $\uparrow$} & \multirow{2}{*}{FID$\downarrow$} & \multirow{2}{*}{MM Dist$\downarrow$} & \multirow{2}{*}{Diversity$\uparrow$} & \multirow{2}{*}{MModality$\uparrow$} \\
\\ \midrule
Real &
$0.511^{\pm.003}$ &
$0.703^{\pm.003}$ &
$0.797^{\pm.002}$ &
$0.002^{\pm.000}$ &
$2.974^{\pm.008}$ &
$9.503^{\pm.065}$ &
- \\ \midrule
Seq2Seq \cite{lin:vigil18} &
$0.180^{\pm.002}$ &
$0.300^{\pm.002}$ &
$0.396^{\pm.002}$ &
$11.75^{\pm.035}$ &
$5.529^{\pm.007}$ &
$6.223^{\pm.061}$ &
- \\
L2P \cite{ahuja2019language2pose}&
$0.246^{\pm.002}$ &
$0.387^{\pm.002}$ &
$0.486^{\pm.002}$ &
$11.02^{\pm.046}$ &
$5.296^{\pm.008}$ &
$7.676^{\pm.058}$ &
- \\
T2G\cite{bhattacharya2021text2gestures} &
$0.165^{\pm.001}$ &
$0.267^{\pm.002}$ &
$0.345^{\pm.002}$ &
$7.664^{\pm.030}$ &
$6.030^{\pm.008}$ &
$6.409^{\pm.071}$ &
- \\
Hier \cite{ghosh2021synthesis}&
$0.301^{\pm.002}$ &
$0.425^{\pm.002}$ &
$0.552^{\pm.004}$ &
$6.532^{\pm.024}$ &
$5.012^{\pm.018}$ &
$8.332^{\pm.042}$ &
- \\
MoCoGAN \cite{tulyakov2018mocogan}&
$0.037^{\pm.000}$ &
$0.072^{\pm.001}$ &
$0.106^{\pm.001}$ &
$94.41^{\pm.021}$ &
$9.643^{\pm.006}$ &
$0.462^{\pm.008}$ &
$0.019^{\pm.000}$ \\
Dance2Music \cite{lee2019dancing}&
$0.033^{\pm.000}$ &
$0.065^{\pm.001}$ &
$0.097^{\pm.001}$ &
$66.98^{\pm.016}$ &
$8.116^{\pm.006}$ &
$0.725^{\pm.011}$ &
$0.043^{\pm.001}$ \\
TEMOS \cite{petrovich2022temos}&
$0.424^{\pm.002}$ &
$0.612^{\pm.002}$ &
$0.722^{\pm.002}$ &
$3.734^{\pm.028}$ &
$3.703^{\pm.008}$ &
$8.973^{\pm.071}$ &
$0.368^{\pm.018}$ \\
Temporal VAE \cite{guo2022generating}&
$0.455^{\pm.003}$ &
$0.636^{\pm.003}$ &
$0.740^{\pm.003}$ &
$1.067^{\pm.002}$ &
$3.340^{\pm.008}$ &
$9.188^{\pm.002}$ &
$2.090^{\pm.083}$ \\
TM2T \cite{guo2022tm2t}&
$0.424^{\pm.003}$ &
$0.618^{\pm.003}$ &
$0.729^{\pm.002}$ &
$1.501^{\pm.017}$ &
$3.467^{\pm.011}$ &
$8.589^{\pm.076}$ &
$\textcolor{blue}{2.424}^{\pm.093}$ \\
MotionDiffuse \cite{zhang2022motiondiffuse}&
$\textcolor{blue}{0.491}^{\pm.001}$ &
$\textcolor{blue}{0.681}^{\pm.001}$ &
$\textcolor{blue}{0.782}^{\pm.001}$ &
$0.630^{\pm.001}$ &
$\textcolor{blue}{3.113}^{\pm.001}$ &
$9.410^{\pm.049}$ &
$1.553^{\pm.042}$ \\
MDM \cite{tevet2022human}&
$0.320^{\pm.005}$ &
$0.498^{\pm.004}$ &
$0.611^{\pm.007}$ &
$0.544^{\pm.044}$ &
$5.566^{\pm.027}$ &
$\textcolor{blue}{9.559}^{\pm.086}$ &
$\textcolor{red}{2.799}^{\pm.072}$ \\
T2M-GPT \cite{zhang2023t2m}&
$0.491^{\pm.003}$ &
$0.680^{\pm.003}$ &
$0.775^{\pm.002}$ &
$\textcolor{red}{0.116}^{\pm.004}$ &
$3.118^{\pm.011}$ &
$\textcolor{red}{9.761}^{\pm.081}$ &
$1.856^{\pm.011}$ \\
\midrule
\midrule
Fg-T2M  &
$\textcolor{red}{0.492}^{\pm.002}$ &
$\textcolor{red}{0.683}^{\pm.003}$ &
$\textcolor{red}{0.783}^{\pm.002}$ &
$\textcolor{blue}{0.243}^{\pm.019}$ &
$\textcolor{red}{3.109}^{\pm.007}$ &
$9.278^{\pm.072}$ &
$1.614^{\pm.049}$ \\
\bottomrule
\end{tabular}%
}
\vspace{-6pt}
\caption{Comparison of text-conditional motion synthesis on HumanML3D~\cite{guo2022generating} dataset. \textcolor{red}{Red} and \textcolor{blue}{Blue} indicate the best and the second-best result, respectively.}
\label{compare_humanml3d}
\end{table}

\begin{table}[t]
\small 
\centering
\resizebox{\columnwidth}{!}{%
\begin{tabular}{@{}lccccccc@{}}
\toprule
\multirow{2}{*}{Methods} & \multirow{2}{*}{R-TOP1 $\uparrow$} & \multirow{2}{*}{R-TOP2 $\uparrow$} & \multirow{2}{*}{R-TOP3 $\uparrow$} & \multirow{2}{*}{FID$\downarrow$} & \multirow{2}{*}{MM Dist$\downarrow$} & \multirow{2}{*}{Diversity$\uparrow$} & \multirow{2}{*}{MModality$\uparrow$} \\
\\ \midrule
Real &
$0.424^{\pm.005}$ &
$0.649^{\pm.006}$ &
$0.779^{\pm.006}$ &
$0.031^{\pm.004}$ &
$2.788^{\pm.012}$ &
$11.08^{\pm.097}$ &
- \\ \midrule
Seq2Seq\cite{lin:vigil18} &
$0.103^{\pm.003}$ &
$0.178^{\pm.005}$ &
$0.241^{\pm.006}$ &
$24.86^{\pm.348}$ &
$7.960^{\pm.031}$ &
$6.744^{\pm.106}$ &
- \\
T2G\cite{bhattacharya2021text2gestures} &
$0.156^{\pm.004}$ &
$0.255^{\pm.004}$ &
$0.338^{\pm.005}$ &
$12.12^{\pm.183}$ &
$6.964^{\pm.029}$ &
$9.334^{\pm.079}$ &
- \\
L2P \cite{ahuja2019language2pose}&
$0.221^{\pm.005}$ &
$0.373^{\pm.004}$ &
$0.483^{\pm.005}$ &
$6.545^{\pm.072}$ &
$5.147^{\pm.030}$ &
$9.073^{\pm.100}$ &
- \\
Hier \cite{ghosh2021synthesis}&
$0.255^{\pm.006}$ &
$0.432^{\pm.007}$ &
$0.531^{\pm.007}$ &
$5.203^{\pm.107}$ &
$4.986^{\pm.027}$ &
$9.563^{\pm.072}$ &
-\\
MoCoGAN \cite{tulyakov2018mocogan}&
$0.022^{\pm.002}$ &
$0.042^{\pm.003}$ &
$0.063^{\pm.003}$ & 
$82.69^{\pm.242}$ & 
$10.47^{\pm.012}$ & 
$3.091^{\pm.043}$ & 
$0.250^{\pm.009}$ \\
Dance2Music \cite{lee2019dancing}&
$0.031^{\pm.002}$ &
$0.058^{\pm.002}$ &
$0.086^{\pm.003}$ & 
$115.4^{\pm.240}$ & 
$10.40^{\pm.016}$ & 
$0.241^{\pm.004}$ & 
$0.062^{\pm.002}$ \\
TEMOS \cite{petrovich2022temos}&
$0.353^{\pm.006}$ &
$0.561^{\pm.007}$ &
$0.687^{\pm.005}$ & 
$3.717^{\pm.051}$ & 
$3.417^{\pm.019}$ & 
$10.84^{\pm.100}$ & 
$0.532^{\pm.034}$ \\
Temporal VAE \cite{guo2022generating}&
$0.361^{\pm.006}$ &
$0.559^{\pm.007}$ &
$0.693^{\pm.007}$ &
$2.770^{\pm.109}$ &
$3.401^{\pm.008}$ &
$10.91^{\pm.119}$ &
$1.482^{\pm.065}$ \\
TM2T \cite{guo2022tm2t}&
$0.280^{\pm.005}$ &
$0.463^{\pm.006}$ &
$0.587^{\pm.005}$ &
$3.599^{\pm.153}$ &
$4.591^{\pm.026}$ &
$9.473^{\pm.117}$ &
$\textcolor{red}{3.292}^{\pm.081}$ \\
MotionDiffuse \cite{zhang2022motiondiffuse}&
$0.417^{\pm.004}$ &
$0.621^{\pm.004}$ &
$0.739^{\pm.004}$ &
$1.954^{\pm.062}$ &
$\textcolor{red}{2.958}^{\pm.005}$ &
$\textcolor{red}{11.10}^{\pm.143}$ &
$0.730^{\pm.013}$ \\
MDM \cite{tevet2022human}&
$0.164^{\pm.004}$ &
$0.291^{\pm.004}$ &
$0.396^{\pm.004}$ &
$\textcolor{red}{0.497}^{\pm.021}$ &
$9.19^{\pm.022}$ &
$10.847^{\pm.109}$ &
$\textcolor{blue}{1.907}^{\pm.214}$ \\
T2M-GPT \cite{zhang2023t2m}&
$0.416^{\pm.006}$ &
$\textcolor{red}{0.627}^{\pm.006}$ &
$\textcolor{blue}{0.745}^{\pm.006}$ &
$\textcolor{blue}{0.514}^{\pm.029}$ &
$\textcolor{blue}{3.007}^{\pm.023}$ &
$10.92^{\pm.108}$ &
$1.570^{\pm.039}$ \\
\midrule
\midrule
Fg-T2M&
$\textcolor{red}{0.418}^{\pm.005}$ &
$\textcolor{blue}{0.626}^{\pm.004}$ &
$\textcolor{red}{0.745}^{\pm.004}$ &
$0.571^{\pm.047}$ &
$3.114^{\pm.015}$ &
$\textcolor{blue}{10.93}^{\pm.083}$ &
$1.019^{\pm.029}$ \\
\bottomrule
\end{tabular}%
}
\vspace{-6pt}
\caption{Comparison of text-conditional motion synthesis on KIT~\cite{plappert2016kit} dataset. \textcolor{red}{Red} and \textcolor{blue}{Blue} indicate the best and the second best result, respectively.}
\label{compare_kit}
\end{table}

Our Fg-T2M framework is evaluated on text-driven motion generation task in this section. We first describe the dataset used and the evaluation metrics in section \ref{sec:Datasets and Evaluation Metrics}. In section \ref{sec:Implementation Details}, we provide implementation details. We compare our framework with the current state-of-the-art methods in section  \ref{sec:Comparison with State-of-the-arts}. Finally, we present qualitative results and visualization for comparison in section \ref{sec:Qualitative Results}.

\subsection{Datasets and Evaluation Metrics}
\label{sec:Datasets and Evaluation Metrics}

\begin{figure*}[t]
    \centering
     \includegraphics[width=0.95\linewidth]{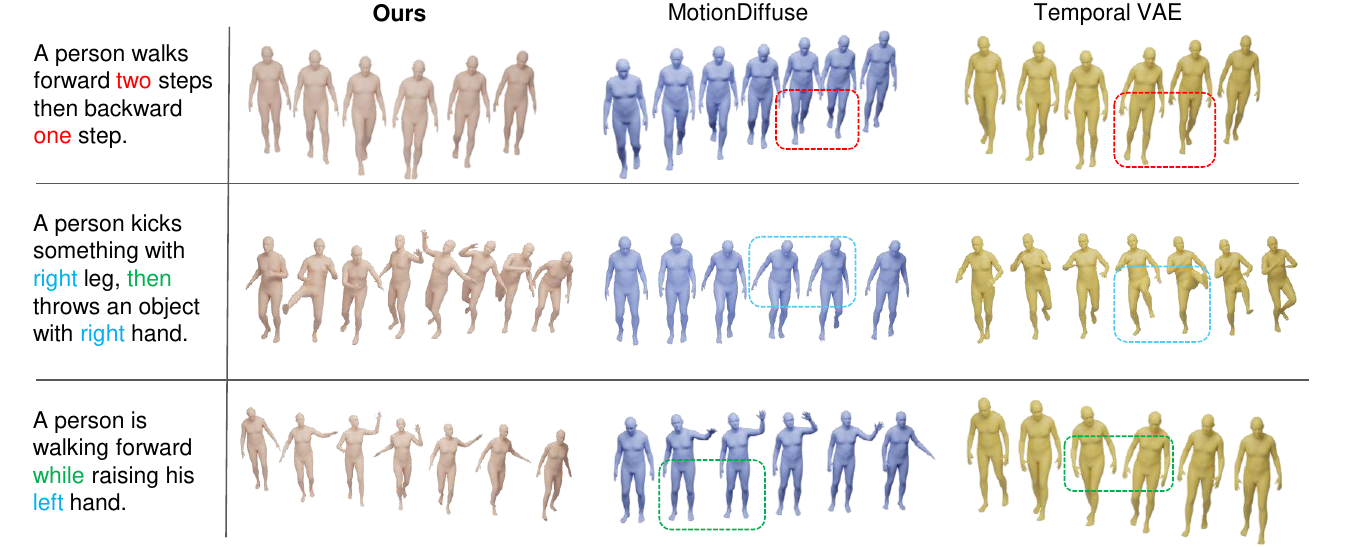}
    \caption{\textbf{Qualitative results}: Our method is compared with two state-of-the-art methods: MotionDiffuse~\cite{zhang2022motiondiffuse} and Temporal VAE~\cite{guo2022generating}. Motion frames are ordered from left to right. Those not matching with text prompt are marked with a box.}
    \label{fig:compare}
\end{figure*}

Several datasets exist for conditional motion generation, such as proposed in \cite{guo2022generating,guo2020action2motion,ji2018large,plappert2016kit}. However, datasets such as \cite{guo2020action2motion} and \cite{ji2018large} are based on action categories and do not provide complete text sentences as conditioning inputs, making them unsuitable for our method. Instead, we use text-driven datasets, 
specifically, \textbf{HumanML3D} dataset~\cite{guo2022generating} and \textbf{KIT} Motion-Language dataset~\cite{plappert2016kit}, for our experiments.

The HumanML3D dataset~\cite{guo2022generating} is a combination of the HumanAct12~\cite{guo2020action2motion} and AMASS~\cite{mahmood2019amass} datasets, comprising 14,616 motions and 44,970 text descriptions across various human activities, such as daily activities, sports, acrobatics, etc., with a total duration of about 28.59 hours. The KIT Motion-Language dataset~\cite{plappert2016kit} consists of 3,911 motion sequences and 6,353 natural language descriptions, with a total duration of around 10.33 hours.

\textbf{Evaluation Metrics} are followed~\cite{guo2022generating}. (1) \textit{R-precision.} For each inferred text-motion pair, 31 mismatched descriptions are randomly selected from the test set. The average top-k precision is obtained by calculating and ranking the Euclidean distance between the motion and each of the 32 descriptions. (2) \textit{Frechet Inception Distance (FID).} FID measures the similarity between the feature distributions extracted from the generated motions and ground truth motions. (3) \textit{Multi-Modal Distance.} The multimodal distance is computed between the text feature and the relevant generated motion feature, concerning the given description. (4) \textit{Diversity.} Diversity evaluates the dissimilarities among all generated motions across all descriptions by computing the mean pairwise Euclidean distance between randomly partitioned groups of motions. (5) \textit{Multimodality.} For a given text description, 32 motion sequences are generated randomly, and multimodality quantifies the dissimilarities among these generated motion sequences. We primarily value R-precision and FID as pivotal performance metrics, which serve as important measures for evaluating the overall quality of generated motions.

\subsection{Implementation Details}
\label{sec:Implementation Details}

The diffusion model uses 1000 diffusion steps and a linearly varying variance $\beta_t$ ranging from 0.0001 to 0.02. LSAM employs a 3-layer GAT network with a corresponding CAPR layer also set to 3. Hyper-parameter $\lambda$ in CAPR is 0.1. Training is performed with the Adam optimizer using a fixed learning rate of 5e-5, a batch size of 128, and NVIDIA GeForce RTX 3090 hardware. The KIT dataset is trained for approximately 40K iterations, while the HumanML3D dataset is trained for about 80K iterations.

\subsection{Comparison with State-of-the-arts}
\label{sec:Comparison with State-of-the-arts}

We compared our method with several state-of-the-art models, including Lin \etal~\cite{lin:vigil18}, Language2Pose~\cite{ahuja2019language2pose}, Ghosh \etal~\cite{ghosh2021synthesis}, MoCoGAN~\cite{tulyakov2018mocogan}, Dance2Music~\cite{lee2019dancing}, TEMOS~\cite{petrovich2022temos}, TM2T~\cite{guo2022tm2t}, Text2Gesture~\cite{bhattacharya2021text2gestures}, Guo \etal~\cite{guo2022generating}, MotionDiffuse~\cite{zhang2022motiondiffuse}, and MDM~\cite{tevet2022human}. Quantitative comparisons of our method with these models on the HumanML3D~\cite{guo2022generating} and KIT~\cite{plappert2016kit} datasets are shown in Tables \ref{compare_humanml3d} and \ref{compare_kit}, respectively. 

Our method achieves competitive performance between text and motion features, as measured by MM Dist, to state-of-the-art methods, while exhibiting significantly higher scores in R-precision and FID. This demonstrates the ability of our method to generate high-quality motions that align with the text prompts. On the other hand, other approaches showcase remarkable competitiveness in diversity and multimodality. However, these aspects should be grounded in accuracy (R-precision) and precision (FID, MMDist) to strengthen their persuasiveness. Otherwise, the diversity or multimodality would be rendered meaningless if the generated motion fails to align with the desired outcome. Therefore, based on our experiments, our method has achieved advanced experimental results and demonstrates robustness in terms of model performance on the two datasets. 

\begin{figure}[t]
    \centering
     \includegraphics[width=0.9\linewidth]{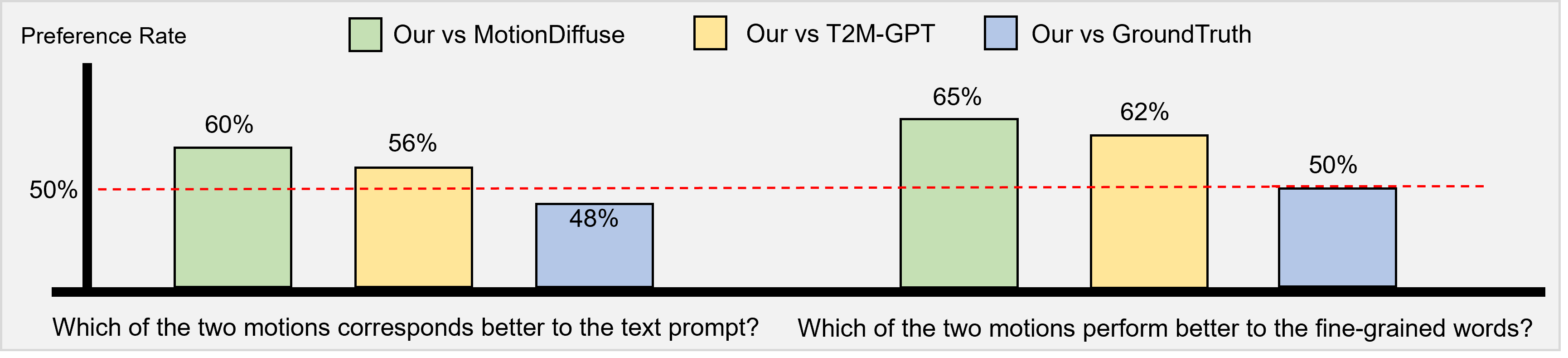}
    \caption{The result of user study.}
    \label{user_study}
\end{figure}

Meanwhile, we design two experiments to evaluate fine-grained control. The first one is conducted on a Harder-HumanML3D dataset. We compress the HumanML3D test set of 4382 data into a Harder-HumanML3D set with 2582 data, by searching for sentences that contain more fine-grained words, like  ``left'',  ``right'', and so on. The comparison results are shown in Table ~\ref{ablation}. Our method exhibits significantly higher scores, indicating a better ability to capture fine-grained details. The second one is a user study in which we collect user preferences with T2M-GPT~\cite{zhang2023t2m} and MotionDiffuse~\cite{zhang2022motiondiffuse}. The statistics of the user study are shown in Figure \ref{user_study}.  Compared with others, our method achieves superior performance in R-Precision, MM Dist, and yields competitive results in FID, which generates motions with comparable quality. Furthermore, for both questions, especially in the fine-grained aspect, ours is preferred over others and even competitive to the ground truth motions.

\begin{table}[t]
\centering
\resizebox{0.9\columnwidth}{!}{%
\begin{tabular}{@{}lccccc@{}}
\toprule
   \multirow{2}{*}{Methods}  & \multicolumn{3}{c}{R-Precision $\uparrow$} & \multirow{2}{*}{FID $\downarrow$} \\
    ~ & Top-1 & Top-2 & Top-3 &\\

    \midrule

MotionDiffuse~\cite{zhang2022motiondiffuse} & $0.439^{\pm.006}$  & $0.614^{\pm.004}$  & $0.725^{\pm.005}$& $0.732^{\pm.021}$ \\

Fg-T2M & $\boldsymbol{0.460}^{\pm.005}$  & $\boldsymbol{0.660}^{\pm.004}$  & $\boldsymbol{0.763}^{\pm.004}$& $\boldsymbol{0.421}^{\pm.013}$ \\
\bottomrule
\end{tabular}%
}

\caption{Experiment results on Harder-HumanML3D split from the HumanML3D~\cite{guo2022generating} test set. }

\label{ablation}
\end{table}
We present ablation results in Table \ref{compare_ablation} to further understand the role of LSAM and CAPR in our method. The results demonstrate that models without the CAPR and LSAM modules exhibit performance degradation. And we also show the comparison of some qualitative examples on the ablation study of this two modules. Additionally, we experiment with the impact of different GAT semantic layers of text information. Since a common dependency tree of motion-text prompt often has three or four depths, a three-layer GAT is sufficient to capture overall nodes information. Hence, a shallow or deep GAT layer hinders the global or local semantic comprehension, resulting in diminished results. We further evaluate the proposed two parts in CAPR. The results show that without CAPR-2 part remarkably reduces the results, which also reflects the importance of fine-grained words for T2M tasks. Finally, the hyper-parameter $\lambda$ in CAPR controls the degree of text information fusion. A larger $\lambda$ dilutes the inherent properties of the motion features, leading to a decrease in precision.

\begin{table}[t]
\centering
\resizebox{0.8\columnwidth}{!}{%
\begin{tabular}{@{}lccccc@{}}
\toprule
\multirow{2}{*}{Methods} & \multirow{2}{*}{R-TOP3 $\uparrow$} & \multirow{2}{*}{FID$\downarrow$} \\
\\ \midrule
Fg-T2M &
$\boldsymbol{0.745}^{\pm.004}$ &
$\boldsymbol{0.571}^{\pm.047}$ \\
\midrule
Fg-T2M (w/o LSAM) & $0.722^{\pm.005}$ & $1.077^{\pm.101}$ \\
Fg-T2M (w/o CAPR) & $0.727^{\pm.007}$  & $0.943^{\pm.089}$  \\
\midrule
Fg-T2M (one layer) & $0.729^{\pm.006}$ & $0.951^{\pm.092}$  \\ 
Fg-T2M (two layers) & $0.738^{\pm.003}$ & $0.692^{\pm.086}$  \\ 
Fg-T2M (four layers) & $0.740^{\pm.004}$ & $0.636^{\pm.102}$  \\
\midrule
Fg-T2M(w/o CAPR-1)   & $0.739^{\pm.006}$& $0.652^{\pm.035}$\\
Fg-T2M(w/o CAPR-2)  & $0.732^{\pm.004}$& $0.764^{\pm.039}$\\
\midrule
Fg-T2M ($\lambda = 0.2$) & $0.738^{\pm.008}$ & $0.686^{\pm.062}$  \\ 
Fg-T2M ($\lambda = 0.3$) & $0.732^{\pm.006}$ & $0.830^{\pm.074}$  \\ 
Fg-T2M ($\lambda = 0.5$) & $0.729^{\pm.011}$ & $1.086^{\pm.104}$  \\
\bottomrule
\end{tabular}%
}
\caption{Ablation analysis on KIT~\cite{plappert2016kit}.  ``layers'' indicates the number of layers of GAT in LSAM. ``(w/o) CAPR-1'' means without multi-modal sentence-level feature-fusion. ``(w/o) CAPR-2'' means without multi-head word-level cross-attention. ``$\lambda$'' is the hyper-parameter of Multi-Modal Sentence-Level Feature-Fusion module in CAPR.}
\label{compare_ablation}
\end{table}





\subsection{Qualitative Results}
\label{sec:Qualitative Results}

Visual results on the HumanML3D~\cite{guo2022generating} dataset are presented in Figure~\ref{fig:compare}, where our method is compared with the state-of-the-art models of MotionDiffuse~\cite{zhang2022motiondiffuse} and Temporal VAE~\cite{guo2022generating}. As can be seen from the examples in the figure, our method generates human motions that more accurately reflect the text prompts. In contrast, the methods of MotionDiffuse~\cite{zhang2022motiondiffuse} and Temporal VAE~\cite{guo2022generating} often result in unrealistic movements. Specifically, these methods tend to only understand one motion within a long text prompt and are not sensitive to specific numerical values, such as 'one' or 'two'. 
As depicted in example one, MotionDiffuse \cite{zhang2022motiondiffuse} only performed a backward motion without considering the required number of steps. On the other hand, Temporal VAE \cite{guo2022generating} successfully achieved the desired effect of walking forward and then backward, but its backward motion step count did not meet the expected requirement.
Moreover, They also make errors in spatial orientation, such as 'left' or 'right', and lack a thorough understanding of temporal issues, such as 'while' and 'then', related to movement. Overall, our proposed method outperforms these models, especially in terms of the issues mentioned above. More diverse samples are presented in the supplementary material.

\section{Conclusion}

We present a novel method for text-driven human motion generation using the diffusion model, which offers several advantages over existing techniques. Specifically, our method leverages two key modules - a linguistics-structure assisted module and a context-aware progressive reasoning module - to effectively model fine-grained words in the text. The former module extracts dependency parsing relationships in the text, while the latter performs hierarchical effective information feature fusion based on graph neural networks. Our quantitative and qualitative results demonstrate that our method outperforms existing techniques in text-driven motion generation tasks. 


\noindent \textbf{Acknowledgements.} This work was supported by the National Natural Science Foundation of China (Project Number: 62272019).

{\small
\bibliographystyle{ieee_fullname}
\bibliography{egbib}
}

\end{document}